\pdfoutput=1

\documentclass[11pt]{article}

\usepackage{booktabs}
\usepackage{multirow}
\usepackage{float}
\usepackage[normalem]{ulem}
\usepackage{caption}

\usepackage{adjustbox}
\usepackage{longtable}
\usepackage{array}

\usepackage{graphicx}

\usepackage{pdflscape}
\usepackage[preprint]{acl}

\usepackage{times}
\usepackage{latexsym}

\usepackage[T1]{fontenc}

\usepackage[utf8]{inputenc}

\usepackage{microtype}

\usepackage{inconsolata}

\usepackage{fontawesome5}

\usepackage{pgfplots}
\pgfplotsset{compat=1.18}

\usepackage[all]{nowidow}

\usepackage{enumitem}

\usepackage{easyReview}

\usepackage{colortbl}
\usepackage{xcolor}
\newcommand{\up}[1]{\textcolor{green!60!black}{~$\uparrow$\tiny #1}}
\newcommand{\down}[1]{\textcolor{red!80!black}{~$\downarrow$\tiny #1}}
\definecolor{colorBase}{HTML}{1F77B4} 
\definecolor{colorFT}{HTML}{FF7F0E}   
\definecolor{color8}{HTML}{2CA02C}    
\definecolor{color6}{HTML}{D62728}    
\definecolor{color4}{HTML}{9467BD}    

\usepackage{hyphenat}

\title{
    \raisebox{-0.2\height}{\includegraphics[width=0.05\textwidth]{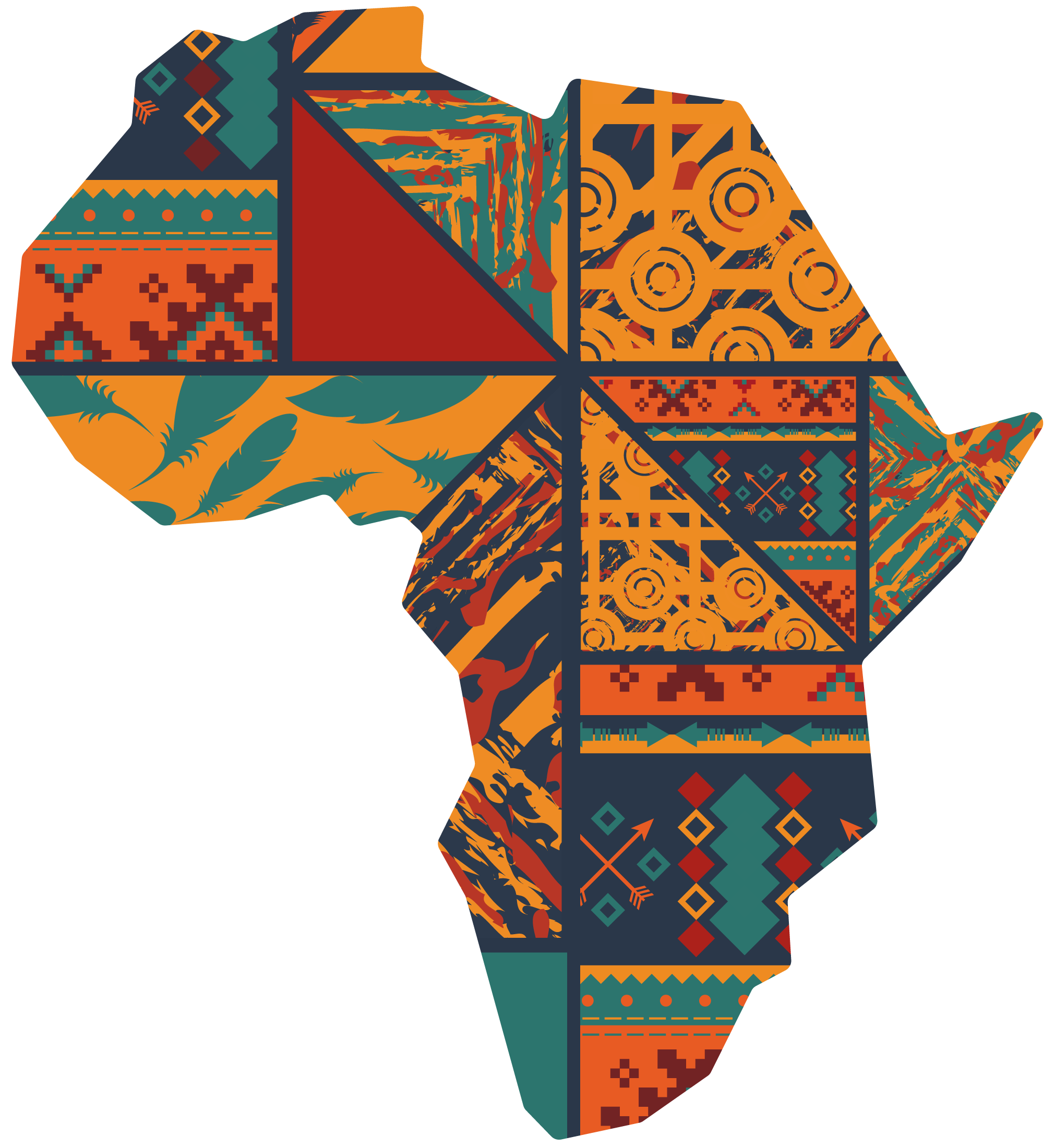}}
    AfriNLLB: Efficient Translation Models for African Languages
}

\author{Yasmin Moslem\Thanks{Equal contribution} \\
    {\normalsize ADAPT Centre} \\
    {\normalsize Trinity College Dublin} \\
    {\normalsize Dublin, Ireland} \\
    {\footnotesize yasmin.moslem@adaptcentre.ie}
    \\\And
    Aman Kassahun Wassie\footnotemark[1]\hspace{8pt} \\
    {\normalsize African Institute for}\hspace{8pt} \\
    {\normalsize Mathematical Sciences (AIMS)}\hspace{8pt} \\
    {\normalsize Addis Ababa, Ethiopia}\hspace{8pt} \\
    {\footnotesize awassie@aimsammi.org\hspace{8pt}}
    \\\And
    Amanuel Gizachew Abebe\footnotemark[1] \\
    {\normalsize Shaggar Institute of} \\
    {\normalsize Technology (SIT)}\\
    {\normalsize Shaggar city, Ethiopia} \\
    {\footnotesize amanuel.g.abebe1@gmail.com}
  }

\begin{document}
\maketitle

\begin{abstract}
\nohyphens{
In this work, we present AfriNLLB, a series of lightweight models for efficient translation from and into African languages. AfriNLLB supports 15 language pairs (30 translation directions), including Swahili, Hausa, Yoruba, Amharic, Somali, Zulu, Lingala, Afrikaans, Wolof, and Egyptian Arabic, as well as other African Union official languages such as Arabic (MSA), French, Portuguese, and Spanish. Our training data covers bidirectional translation between English and 13 languages, and between French and two languages (Lingala and Wolof). 

AfriNLLB models are based on NLLB-200 600M, which we compress using iterative layer pruning and quantization. We fine-tune the pruned models on parallel corpora we curated for African languages, employing knowledge distillation from a larger teacher model. Our work aims at enabling efficient deployment of translation models for African languages in resource-constrained settings. 

Our evaluation results demonstrate that AfriNLLB models achieve performance comparable to the baseline while being significantly faster. We release two versions of the AfriNLLB models, a Transformers version that allows further fine-tuning and a CTranslate2 version for efficient inference. Moreover, we release all the training data that we used for fine-tuning the baseline and pruned models to facilitate further research.
}

\end{abstract}

\begin{table*}[ht]
\centering
\begin{small}

\begin{tabular}{@{}llllr@{}}
	\toprule
	\textbf{Family} & \textbf{Subfamily} & \textbf{Name} & \textbf{Code} & \textbf{Regions} \\
\midrule
\multirow{4}{*}{Afro-Asiatic} & Chadic & Hausa & hau\_Latn  & West Africa (Nigeria, Niger) \\
& Cushitic & Somali & som\_Latn  & Horn of Africa (Somalia, Ethiopia, Djibouti, Kenya) \\
& Semitic & Amharic & amh\_Ethi  & Horn of Africa (Ethiopia) \\
& Semitic & Egyptian Arabic & arz\_Arab  & North Africa (Egypt) \\
\midrule
\multirow{1}{*}{Indo-European} & Germanic & Afrikaans & afr\_Latn & Southern Africa (South Africa, Namibia) \\
\midrule
\multirow{5}{*}{Niger-Congo} & Atlantic & Wolof & wol\_Latn  & West Africa (Senegal, Gambia, Mauritania) \\
& Bantu & Lingala & lin\_Latn  & Central Africa (Congo) \\
& Bantu & Swahili & swh\_Latn  & East Africa (Tanzania, Kenya) \\
& Bantu & Zulu & zul\_Latn  & Southern Africa (South Africa) \\
& Volta-Niger & Yoruba & yor\_Latn & West Africa (Nigeria, Benin) \\
\bottomrule
\end{tabular}
\end{small}

\caption{African Languages in AfriNLLB}
\label{tab:languages-african}
\end{table*}

\begin{table*}[ht]
\centering
\begin{small}

\begin{tabular}{@{}llllr@{}}
	\toprule
	\textbf{Family} & \textbf{Subfamily} & \textbf{Name} & \textbf{Code} & \textbf{Regions} \\
\midrule
\multirow{1}{*}{Afro-Asiatic} & Semitic &  Arabic, Modern Standard & arb\_Arab & North Africa (formal use) \\
\midrule
\multirow{4}{*}{Indo-European} & Germanic & English & eng\_Latn  & Southern Africa (South Africa) \\
& Romance & French & fra\_Latn  & Africa-wide (mostly L2) \\
& Romance & Portuguese & por\_Latn  & Southern Africa (Angola, Mozambique) \\
& Romance & Spanish & spa\_Latn  & Central Africa (Equatorial Guinea) \\
\bottomrule
\end{tabular}
\end{small}

\caption{Non-Native Languages in AfriNLLB}
\label{tab:languages-high-resource}
\end{table*}

\section{AfriNLLB: Background \& Motivation}

Africa is a linguistically rich continent, with over 2,000 native languages \cite{Grimes1996-Ethnologue,Heine2000-AfricanLanguages}. Although African languages have millions of native speakers, most of them are low-resource languages \cite{Azime2024-Walia,Wassie2024-GeezMT,Adelani2025-IrokoBench,Farouq2025-Bemba,Ojo2025-AfroBench}. This results in a scarcity of African datasets and models for diverse natural language processing tasks, including machine translation (MT). Since MT resources for African languages are scattered across multiple sources, gathering these resources for fine-tuning open-source models is costly and time-consuming. Moreover, providing translation support for speakers of these low-resource languages in governmental and health sectors remains a significant challenge \cite{Anastasopoulos2020-TICO-19,Wassie2025-DomainSpecificMT}.

\textbf{AfriNLLB} seeks to bridge this gap by delivering efficient translation models and curated training data.%
\footnote{\url{https://github.com/AfriNLP/AfriNLLB}}\textsuperscript{,}\footnote{\url{https://hf.co/collections/AfriNLP/afrinllb}}
Language selection for AfriNLLB considered several factors, including the number of native speakers in Africa and dataset availability. The AfriNLLB models are based on NLLB-200 \cite{NLLB2022}, and support 15 language pairs (30 translation directions), including 10 native African languages: Swahili, Hausa, Yoruba, Amharic, Somali, Zulu, Lingala, Afrikaans, Wolof, and Egyptian Arabic (cf.~Table~\ref{tab:languages-african}). Additionally, we include 5 of the official languages of the African Union, namely Arabic (MSA), English, French, Portuguese, and Spanish. Since several African languages share some lexicon with these languages due to historical contact, multilingual models can leverage this linguistic overlap through transfer learning from high-resource languages to enhance the performance of low-resource languages \cite{Liu2020-mBART,Fan2021-M2M}.

AfriNLLB is a series of efficient multilingual open-source models for African languages, motivated by multiple goals:

\begin{itemize}[itemsep=3pt,topsep=4pt,parsep=0pt]
    \item Gathering and curating bilingual training datasets for African languages
    \item Building lightweight MT models specialized in translating African languages, utilizing compression approaches such as pruning and quantization
    \item Open-sourcing the code, training data, and models we have created
    \item Sharing our approaches and lessons learned to facilitate future work in this area
\end{itemize}

\section{Data}
\label{sec:data}

We employ multi-stage fine-tuning before and after model pruning. First, we fine-tune the baseline NLLB-200 600M to improve the performance for African languages. Afterwards, we fine-tune the pruned models again to restore the translation performance. For this purpose, we collect datasets primarily in African languages (Swahili, Hausa, Yoruba, Amharic, Somali, Zulu, Lingala, Afrikaans, Wolof, and Egyptian Arabic) and a few relevant high-resource languages (Arabic (MSA), French, Spanish, Portuguese). 

\subsection{Data Sources}

We mainly collect the datasets from OPUS \cite{Tiedemann2012-OPUS} and Hugging Face \cite{Lhoest2021-Datasets}, with additional data from GitHub and other publicly available online sources. This results in a total of 1.2M samples for 11 African language pairs (9 from/into English, and 2 from/into French). For high-resource languages (Arabic, French, Spanish, Portuguese), we focus on collecting only 1.5M for processing, filter the data, and then sample 200k from each language pair for training. Table~\ref{tab:data-summary} summarizes data before and after filtering, while Table~\ref{tab:data-details} elaborates on data sources.

\subsection{Data Processing}

To ensure the quality of data, we process the datasets in a four-stage pipeline: (i) rule-based filtering, (ii) language detection, (iii) semantic filtering, and (iv) quality estimation. While rule-based filtering uses predefined rules, the other pipeline stages employ a model to generate scores and filter the data based on a threshold. We experimented with different threshold values and found 0.6 to be a reasonable choice. 

\paragraph{Rule-based filtering} involves deduplication, dropping empty segments, and removing HTML tags. We also filter out sentence pairs with lengths less than 3 or greater than 200 characters. Moreover, to avoid misaligned segments, we remove translation pairs exceeding the 2x source-target length ratio.

\paragraph{Language detection} discards segments that are unlikely to be in the expected language. We use two language detector models, AfroLID \cite{Adebara2022-AfroLID} for the African languages and fastText \cite{Joulin2017-fastText} for the rest of the languages.

\paragraph{Semantic filtering} evaluates the translation pairs with cosine similarity scores derived from sentence embedding models, using the Sentence-Transformers library \cite{Reimers2019-SentenceTransformers}.
To handle all the languages, we employ different embedding models based on language support. We use \textit{DistilUSE} \cite{Reimers2020-DistilUSE,Yang2020-mUSE} for all high-resource language pairs and \textit{LaBSE} \cite{Feng2022-LaBSE} for African languages. We apply semantic filtering for all languages except Lingala as we could not find an embedding model that supports it.

\paragraph{Quality estimation} is the final stage of the filtering pipeline, in which we apply reference-free evaluation of the translation and exclude segments that are lower than the threshold. We use COMET \cite{Rei2020-COMET} for high-resource language pairs, and Masakhane’s model AfriCOMET-QE-STL \cite{Wang2024-AfriCOMET} for African languages.

After thoroughly processing the dataset, we merge the datasets and deduplicate the combined dataset to avoid repetition from different sources. We ended up with a total of 6.4M. However, to mitigate data imbalance, we downsampled the high-resource languages to only 200k per language pair. This results in a total of 1.6M samples (3.2M bidirectional samples, after reversing the dataset), which we use for training. The dataset size for each language direction is presented in Table~\ref{tab:data-summary}, and elaborated in Table~\ref{tab:data-details}.

\subsection{Validation and Test Data} 
We use Flores200\footnote{\scriptsize\url{https://hf.co/datasets/facebook/flores}} \cite{NLLB2022} for validation and test, as it covers all the languages in our experiments. We use the \textit{dev} split (997 segments) of Flores200 for validation during training, and for layer importance evaluation as part of iterative layer pruning (cf.~Section~\ref{sec:experiments}), and use the \textit{devtest} split (1,012 segments) for testing and evaluation of our models.

\begin{table}[htbp]
\centering
\begin{scriptsize}
\begin{tabular}{llrrr}
\toprule
\multicolumn{2}{c}{\textbf{Language Pair}} 
  & \textbf{Initial} 
  & \textbf{Processed} 
  & \textbf{Sampled} \\
\midrule

\multirow{11}{*}{eng\_Latn}
  & afr\_Latn & 192,541 & 161,644 & 161,644 \\
  & amh\_Ethi & 156,739 & 85,010 & 85,010 \\
  & arz\_Arab & 85,942 & 84,170 & 84,170 \\
  & hau\_Latn & 222,387 & 155,881 & 155,881 \\
  & som\_Latn & 87,521 & 43,657 & 43,657 \\
  & swh\_Latn & 286,687 & 181,045 & 181,045 \\
  & wol\_Latn & 34,956 & 31,170 & 31,170 \\
  & yor\_Latn & 34,720 & 22,626 & 22,626 \\
  & zul\_Latn & 38,532 & 33,189 & 33,189 \\

& arb\_Arab & 1,526,102 & 1,424,237 & 200,000 \\
  & fra\_Latn & 1,500,000 & 1,483,951 & 200,000 \\
  & por\_Latn & 1,500,000 & 1,401,671 & 200,000 \\
  & spa\_Latn & 1,500,000 & 1,324,681 & 200,000 \\
\midrule
\multirow{2}{*}{fra\_Latn}
  & wol\_Latn & 10,745 & 9,071 & 9,071 \\
  & lin\_Latn & 8126 & 1,948 & 1,948 \\
\midrule
  & Total & 7,184,998 & 6,443,951 & 1,609,411 \\
\bottomrule
\end{tabular}
\end{scriptsize}

\caption{Parallel corpus sizes before and after processing from and into English and French.
Since all data is reversed to create the opposite translation direction, the final dataset size is effectively doubled.}
\label{tab:data-summary}
\end{table}

\section{Methodology}
\label{sec:experiments}

In our experiments, we apply iterative layer pruning to the \textit{NLLB-200 600M} model after fine-tuning it on the training dataset. This approach incrementally identifies and removes layers with minimal contribution to translation quality, one layer at a time. The pruned models resulting from this process are then fine-tuned again 
to restore most of the translation quality of the baseline model. The resulting models are smaller and faster while retaining or outperforming the quality of the baseline. The following points elaborate on the process.

\begin{table*}[ht]
\centering
\small
\begin{tabular}{llccc|cc}
\toprule
\textbf{Direction} & \textbf{Model} & \textbf{BLEU ↑} & \textbf{chrF++ ↑} & \textbf{COMET ↑} & \textbf{Throughput (toks/s) ↑} & \textbf{Time (s) ↓} \\
\midrule
\multirow{3}{*}{\textbf{xx-en}} 
& NLLB 600M (Baseline) & 33.81 & 56.22 & 71.11 & \cellcolor{orange!8}{1469.96} & \cellcolor{orange!8}{21.02} \\
& NLLB 600M + FT & 35.15 & 57.61 & 71.87 & \cellcolor{orange!8}{1530.94} & \cellcolor{orange!8}{20.39} \\
\cline{2-7}\addlinespace[4pt]
& Pruned + FT & 34.01 & 56.98 & 71.20 & \cellcolor{green!4}{1807.61} & \cellcolor{green!4}{17.38}  \\

& Pruned + FT (FP16) & 34.05 & 56.99 & 71.19 & \cellcolor{green!15}{3513.32} & \cellcolor{green!15}{8.96}  \\

\midrule

\multirow{3}{*}{\textbf{en-xx}} 
& NLLB 600M (Baseline) & 22.70 & 47.89 & 69.36 & \cellcolor{orange!8}{1530.10} & \cellcolor{orange!8}{28.09}  \\
& NLLB 600M + FT &24.28 & 49.97 & 70.91 & \cellcolor{orange!8}{1610.23} & \cellcolor{orange!8}{26.98} \\
\cline{2-7}\addlinespace[4pt]
& Pruned + FT &  24.17 & 50.05 & 70.37 & \cellcolor{green!4}{1946.61} & \cellcolor{green!4}{22.51}  \\

& Pruned + FT (FP16) & 24.15 & 50.06 & 70.41 & \cellcolor{green!15}{3732.72} & \cellcolor{green!15}{11.98}  \\

\midrule \midrule

\multirow{3}{*}{\textbf{xx-fr}} 
& NLLB 600M (Baseline) & 16.41 & 38.83 & 17.34 & \cellcolor{orange!8}{1475.48} & \cellcolor{orange!8}{26.46} \\
& NLLB 600M + FT & 17.91 & 40.45 & 18.47 & \cellcolor{orange!8}{1524.32} & \cellcolor{orange!8}{26.12} \\
\cline{2-7}\addlinespace[4pt]
& Pruned + FT &  17.43 & 40.21 & 14.52 & \cellcolor{green!4}{1845.09} & \cellcolor{green!4}{21.61} \\

& Pruned + FT (FP16) & 17.38 & 40.18 & 14.53 & \cellcolor{green!15}{3569.23} & \cellcolor{green!15}{11.17}  \\

\midrule

\multirow{3}{*}{\textbf{fr-xx}} 
& NLLB 600M (Baseline) & 9.44 & 33.42 & 19.25 & \cellcolor{orange!8}{1047.18} & \cellcolor{orange!8}{49.92} \\
& NLLB 600M + FT &10.98 & 35.68 & 21.33 & \cellcolor{orange!8}{1081.84} & \cellcolor{orange!8}{51.56} \\
\cline{2-7}\addlinespace[4pt]
& Pruned + FT &  10.20 & 35.21 & 20.04  & \cellcolor{green!4}{1261.66} & \cellcolor{green!4}{49.91} \\

& Pruned + FT (FP16) &  10.11 & 35.13 & 20.03 & \cellcolor{green!15}{2313.85} & \cellcolor{green!15}{31.15} \\

\bottomrule
\end{tabular}
\caption{Average Performance by Translation Direction. The category en$\,\leftrightarrow\,$xx includes 13 language pairs (26 translation directions), while the category fr$\,\leftrightarrow\,$xx includes 2 language pairs for Lingala and Wolof (4 translation directions). The pruned models are up to 20\% faster than the baseline without quantization, and 57\% faster with float16 quantization. While more efficient, the translation quality of the compressed models is comparable with the fine-tuned NLLB-200 model. Table \ref{tab:eval-all} elaborates on the experimental results.}
\label{tab:avg-performance}
\end{table*}

\vspace{0.3cm}

\paragraph{Layer importance evaluation:} We conduct layer importance evaluation by measuring translation performance without each layer. In this greedy layer pruning approach \citep{Peer2022-GreedyLayerPruning,Rostami2024-CULL-MT,Moslem2025-IterativePruning,Moslem2025-EfficientSpeech-IWSLT}, to prune \(n + 1\) layers, only a single optimal layer to prune must be added to the already known solution for pruning $n$ layers. After identifying and removing the least critical layer, we repeat the layer importance evaluation on the remaining layers until reaching our $n$ pruning target. We observe that while removing certain layers of the model (e.g. the first or last layer) substantially degrades translation performance, others result in minimal performance drops. Following \citet{Moslem2025-EfficientSpeech-IWSLT}, we use the chrF++ metric for layer importance evaluation for both better efficiency and quality. We use the dev split of the Flores200 dataset, mainly where African languages are the target, to improve their translation quality. In the future, we plan to experiment with using both directions.

\paragraph{Layer pruning:} We iteratively prune one decoder layer at a time, selecting the layer whose removal has the least negative impact on translation quality, measured by chrF++ scores. At each iteration, we evaluate the translation performance of the pruned model on the dev split of the Flores200 dataset, after removing each candidate layer. The layer whose removal yields the best performance is eventually pruned. This process continues until a predefined number of layers (4, 6, or 8 layers) have been removed. By iteratively removing the least important layers, this performance-guided method produces a more compact model that can be fine-tuned further to recover the translation quality of the original model.
We also experimented with middle layer pruning and found that iterative layer pruning yields better results (cf.~Section \ref{sec:ablation}).

\paragraph{Fine-tuning:} We employ multi-stage fine-tuning. First, we fine-tune the baseline NLLB-200 model on the training dataset to improve its quality for African languages. Since pruning the fine-tuned models results in performance degradation, the pruning step is followed by fine-tuning the pruned model for 1 epoch using the training dataset (cf.~Section~\ref{sec:data}). During training, we use a learning rate of 5e-5, a batch size of 8, gradient accumulation steps of 4, and early stopping with a patience value of 10 evaluation runs. The evaluation takes place every 1000 training steps. The final saved model is the best model based on the evaluation loss score. The training is conducted on one A40 48GB GPU. We use the \textit{Transformers} framework\footnote{\scriptsize\url{https://github.com/huggingface/transformers}} \cite{Wolf2020-Transformers} for training. As illustrated by Table~\ref{tab:avg-performance}, this fine-tuning step successfully recovers the translation quality of the baseline model.

\paragraph{Knowledge distillation:} To improve the quality of our models, we employ sequence-level knowledge distillation \cite{Kim2016-KnowledgeDistillation,Crego2016-KnowledgeDistillation,Gandhi2023-Distil-Whisper}, where the student model is fine-tuned on a combination of authentic data and synthetic data generated by the teacher model for the same training dataset. In this case, the teacher model is the NLLB-200 3.3B baseline, while the students are the NLLB-200 600M baseline and then the pruned models based on our fine-tuned version. After generating the data, we filter it by removing duplicates (exact matches in the target side of the authentic data), and we follow the filtering pipeline we use for processing the original training data (cf.~Section~\ref{sec:data}). The knowledge distillation data after filtering is 568k segments for African languages.

\section{Evaluation and Results}

For inference, we use CTranslate2\footnote{\scriptsize\url{https://github.com/OpenNMT/CTranslate2}} \cite{Klein2020-Efficient}, with a beam size of 3 and a batch size of 1024 tokens, on an A40 48GB GPU.

To evaluate our systems, we calculated BLEU \citep{Papineni2002-BLEU},  chrF++ \citep{Popovic2017-chrF++}, as implemented in the sacreBLEU library\footnote{\scriptsize\url{https://github.com/mjpost/sacrebleu}} \citep{Post2018-sacreBLEU}. For semantic evaluation, we use AfriCOMET \cite{Wang2024-AfriCOMET} for African languages, and COMET \citep{Rei2020-COMET} for Arabic and European languages.\footnote{\scriptsize In particular, we used the \textit{“africomet-mtl”} model for AfriCOMET and the \textit{“wmt22-comet-da”} model for COMET.}

The process of iterative layer pruning of 4 decoder layers created a 548M model that is 23\% faster in average than the baseline. Moreover, the quality degradation caused by pruning has been mitigated through fine-tuning and knowledge distillation. As demonstrated by Table \ref{tab:avg-performance} and elaborated by Table~\ref{tab:eval-details}, by the end of the process, the pruned model could recover most of the translation quality of the baseline model. Moreover, quantization (float16) of the pruned model further enhanced the inference performance, making the model 57\% faster than the baseline.

\subsection{Ablation Study}
\label{sec:ablation}

In this ablation study, we compare three scenarios: (i) removing middle layers%
\footnote{\scriptsize For middle layer pruning, we remove layers 4 to 7 inclusively.}
instead of iteratively determining the layers to remove based on layer importance evaluation (cf.~Section~\ref{sec:experiments}),  (ii) pruning both encoder and decoder layers instead of pruning decoder layers only, and (iii) pruning various values of the decoder layers, namely 4, 6, and 8 layers.

\begin{figure}[ht]
    \centering
    \begin{tikzpicture}
        \begin{axis}[
            width=8.3cm,
            height=11cm,
            xlabel={\scriptsize Inference Throughput (Tokens/sec)},
            ylabel={\scriptsize Translation Quality (Avg chrF++)},
            grid=major,
            grid style={dashed, gray!30},
            legend pos=north west,
            xmin=5, xmax=2600,
            ymin=5, ymax=65,
            tick label style={font=\tiny},
            xtick={500,1000,...,2500},
            ytick={5,10,...,65},
            scatter/classes={
                middle={mark=square*,purple},
                iterative={mark=*,green}
            },
            legend style={
                align=left,
                cells={anchor=west}
            },
        ]
        
        \addplot[
            scatter,
            only marks,
            mark=square,
            mark size=2.0pt,
            color=purple!60!blue,
            nodes near coords,
            point meta=explicit symbolic,
            every node near coord/.append style={anchor=south east, font=\tiny, xshift=0pt, yshift=-6pt}
        ] table [meta=label] {
            x        y       label
            926.72   21.59     {Mid-548}
            1102.88   8.24    {Mid-498}
        };
        \addlegendentry{\tiny Middle Pruned (Mid)}

        \addplot[
            scatter,
            only marks,
            mark=o,
            mark size=2pt,
            color=green!40!black,
            nodes near coords,
            point meta=explicit symbolic,
            every node near coord/.append style={anchor=south east, font=\tiny, xshift=0pt, yshift=-6pt}
        ] table [meta=label] {
            x        y       label
            1582.88     47.68    {Iter-548}
            1240.89     38.64    {Iter-498}
        };
        \addlegendentry{\tiny Iterative Pruned (Iter)}

        \addplot[
            scatter,
            only marks,
            mark=square*,
            mark size=2.0pt,
            color=purple!80!blue,
            nodes near coords,
            point meta=explicit symbolic,
            every node near coord/.append style={anchor=south east, font=\tiny, xshift=0pt, yshift=-6pt},
            forget plot
        ] table [meta=label] {
            x        y       label
            1857.95   50.93   {Mid-FT-548}
        };
        
        \addplot[
            scatter,
            only marks,
            mark=square*,
            mark size=2.0pt,
            color=purple!80!blue,
            nodes near coords,
            point meta=explicit symbolic,
            every node near coord/.append style={anchor=south west, font=\tiny, xshift=0pt, yshift=-6pt}
        ] table [meta=label] {
            x        y       label
            1896.17   48.75   {Mid-FT-498}
        };
        \addlegendentry{\tiny Middle Pruned, Fine-tuned (Mid-FT)}
        

        \addplot[
            scatter,
            only marks,
            mark=*,
            mark size=2pt,
            color=green!70!black,
            nodes near coords,
            point meta=explicit symbolic,
            every node near coord/.append style={font=\tiny, anchor=south east, yshift=2pt, xshift=20pt},
            forget plot
        ] table [meta=label] {
            x        y       label
            1833.95     51.41   {Iter-FT-548}
        };
        
        \addplot[
            scatter,
            only marks,
            mark=*,
            mark size=2pt,
            color=green!70!black,
            nodes near coords,
            point meta=explicit symbolic,
            every node near coord/.append style={font=\tiny, anchor=north west, yshift=6pt}
        ] table [meta=label] {
            x        y       label
            2091.46     50.83   {Iter-FT-498}
        };
        \addlegendentry{\tiny Iterative Pruned, Fine-tuned (Iter-FT)}

        \end{axis}
    \end{tikzpicture}
    
    \caption{Quality-Efficiency Comparison. The iterative-pruned models demonstrate a superior balance of speed and quality compared to the middle-pruned variants. The 548M models include 12 encoder layers and 8 decoder layers (i.e. 4 decoder layers are pruned), while the 498M models include 8 encoder layers and 8 decoder layers (i.e. 8 layers are pruned, 4 from the encoder and 4 from the decoder). The chart reports the average chrF++ scores across all language pairs before and after fine-tuning the pruned models.}
    \label{fig:ablation-study}
\end{figure}
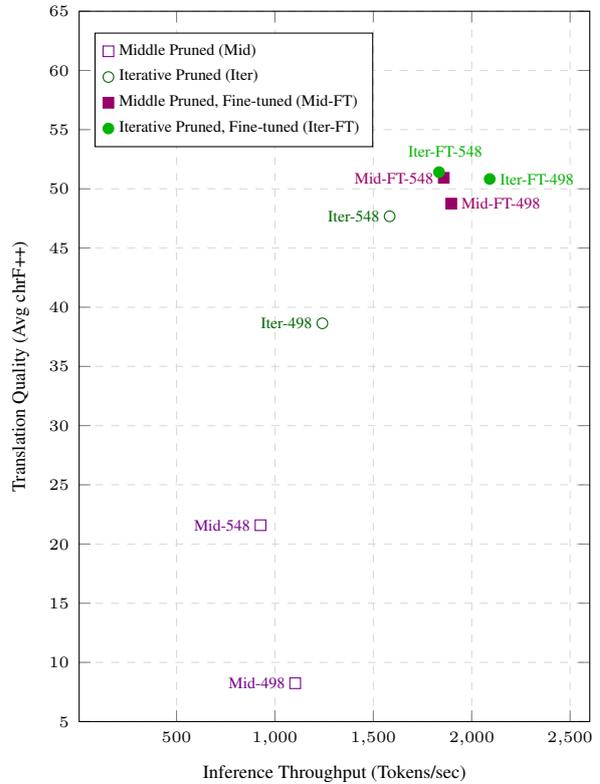

We observe that iterative layer pruning clearly outperforms middle layer pruning in both cases of removing decoder layers only or both encoder and decoder layers. Fine-tuning after pruning is crucial in all cases, as it mitigates the effect of pruning on performance. Figure \ref{fig:ablation-study} illustrates four pruned models, both before and after fine-tuning:

\begin{itemize}[itemsep=3pt,topsep=4pt,parsep=0pt]
    \item {\small Middle pruning, 4 decoder layers (Mid 548M)}
    \item {\small Middle pruning, 4 encoder layers and 4 decoder layers (Mid 498M)}
    \item {\small Iterative pruning, 4 decoder layers (Iter 548M)}
    \item {\small Iterative pruning, 4 encoder layers and 4 decoder layers (Iter 498M)}
\end{itemize}

When it comes to removing encoder layers in addition to decoder layers, it is not clear to what extent this affects the quality. Obviously, removing encoder layers reduces the size of the model further, which can cause performance degradation. Keeping encoder layers intact was recommended by previous work on speech \cite{Gandhi2023-Distil-Whisper,Moslem2025-EfficientSpeech-IWSLT}, which poses the question whether the same concept applies to text-based encoder-decoder models such as NLLB-200. We intend to investigate this further in future work.

\begin{figure*}[htbp]
    \centering
    \begin{tikzpicture}
    \begin{axis}[
        ybar,
        width=\textwidth,
        height=7.3cm,
        title={\textbf{Translation Performance} (English/French $\rightarrow$ African)},
        ylabel={chrF++ Score},
        symbolic x coords={eng-afr, eng-amh, eng-arb, eng-arz, eng-fra, eng-hau, eng-por, eng-som, eng-spa, eng-swh, eng-wol, eng-yor, eng-zul, fra-lin, fra-wol},
        xtick=data,
        x tick label style={rotate=45, anchor=east, font=\footnotesize},
        ymin=20, ymax=82,
        tick label style={font=\scriptsize},
        ytick={20,30,...,80},
        ymajorgrids=true,
        grid style=dashed,
        bar width=2pt,
        legend style={at={(0.5,1.15)}, anchor=south, legend columns=5, font=\footnotesize},
        enlarge x limits=0.04,
        ylabel style={font=\small},
        title style={font=\small},
        nodes near coords,
        nodes near coords style={
            font=\fontsize{5}{6}\selectfont,
            rotate=90, 
            anchor=west, 
            /pgf/number format/fixed, 
            /pgf/number format/precision=1
        },
    ]

    \addplot[fill=colorBase, draw=none] coordinates {
        (eng-afr, 61.76) (eng-amh, 36.96) (eng-arb, 51.30) (eng-arz, 40.81) (eng-fra, 66.61) 
        (eng-hau, 48.99) (eng-por, 67.17) (eng-som, 41.45) (eng-spa, 52.59) (eng-swh, 57.68) 
        (eng-wol, 23.56) (eng-yor, 22.87) (eng-zul, 50.78) (fra-lin, 45.00) (fra-wol, 21.84)
    };

    \addplot[fill=colorFT, draw=none] coordinates {
        (eng-afr, 64.95) (eng-amh, 38.74) (eng-arb, 52.37) (eng-arz, 43.55) (eng-fra, 68.15) 
        (eng-hau, 52.54) (eng-por, 65.27) (eng-som, 41.43) (eng-spa, 52.55) (eng-swh, 61.48) 
        (eng-wol, 27.42) (eng-yor, 28.71) (eng-zul, 52.39) (fra-lin, 46.75) (fra-wol, 24.61)
    };

    \addplot[fill=color8, draw=none] coordinates {
        (eng-afr, 66.23) (eng-amh, 38.37) (eng-arb, 52.51) (eng-arz, 44.43) (eng-fra, 66.99) 
        (eng-hau, 53.21) (eng-por, 65.33) (eng-som, 40.98) (eng-spa, 51.40) (eng-swh, 62.23) 
        (eng-wol, 28.63) (eng-yor, 29.20) (eng-zul, 51.19) (fra-lin, 44.97) (fra-wol, 25.45)
    };

    \addplot[fill=color6, draw=none] coordinates {
        (eng-afr, 65.30) (eng-amh, 37.17) (eng-arb, 51.46) (eng-arz, 43.86) (eng-fra, 66.07) 
        (eng-hau, 53.04) (eng-por, 64.51) (eng-som, 39.76) (eng-spa, 50.50) (eng-swh, 61.38) 
        (eng-wol, 29.65) (eng-yor, 28.95) (eng-zul, 49.73) (fra-lin, 43.04) (fra-wol, 27.24)
    };

    \addplot[fill=color4, draw=none] coordinates {
        (eng-afr, 64.53) (eng-amh, 33.16) (eng-arb, 49.32) (eng-arz, 42.88) (eng-fra, 63.62) 
        (eng-hau, 52.49) (eng-por, 63.27) (eng-som, 37.80) (eng-spa, 49.01) (eng-swh, 60.72) 
        (eng-wol, 28.18) (eng-yor, 28.15) (eng-zul, 48.25) (fra-lin, 38.98) (fra-wol, 25.85)
    };

    \legend{Baseline, FT+KD, 8-dec, 6-dec, 4-dec}

    \end{axis}
    \end{tikzpicture}
    \caption{Translation performance (chrF++) from English/French to African languages.
    }
    \label{fig:eng-afr}
\end{figure*}
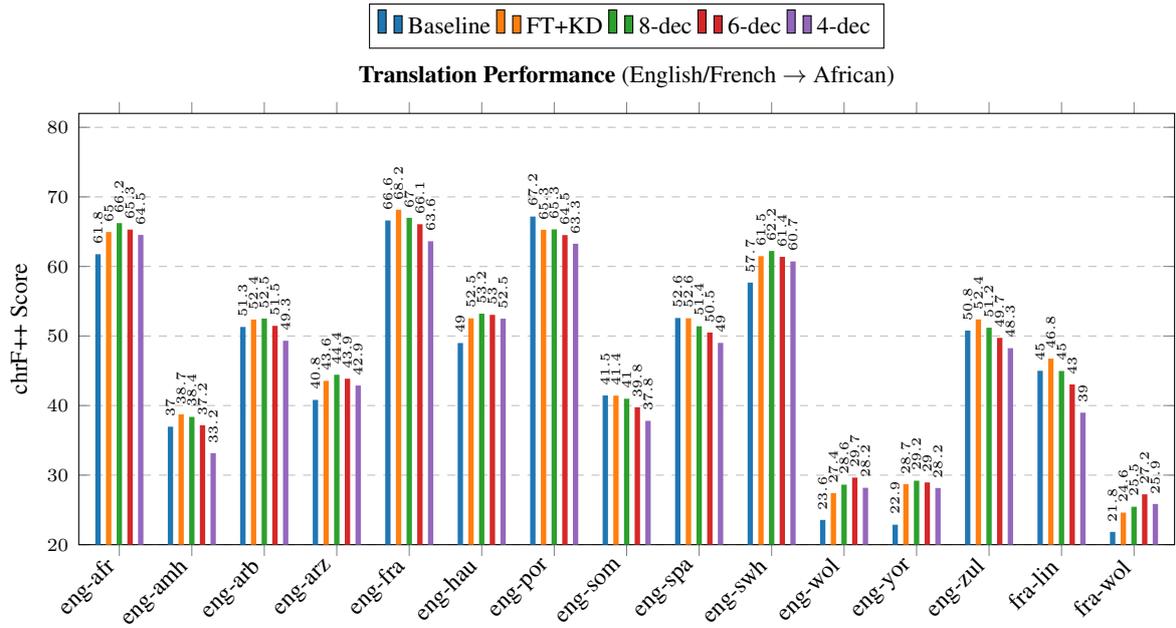

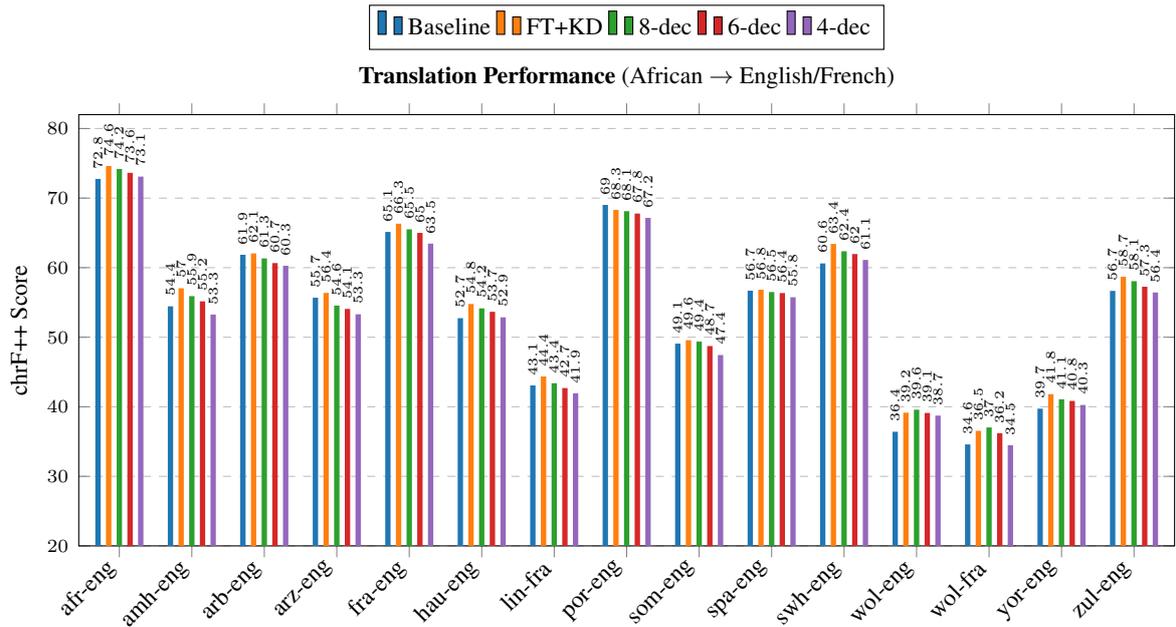
\begin{figure*}[htbp]
    \centering
    \begin{tikzpicture}
    \begin{axis}[
        ybar,
        width=\textwidth,
        height=7.3cm,
        title={\textbf{Translation Performance} (African $\rightarrow$ English/French)},
        ylabel={chrF++ Score},
        symbolic x coords={afr-eng, amh-eng, arb-eng, arz-eng, fra-eng, hau-eng, por-eng, som-eng, spa-eng, swh-eng, wol-eng, yor-eng, zul-eng, lin-fra, wol-fra},
        xtick=data,
        x tick label style={rotate=45, anchor=east, font=\footnotesize},
        ymin=20, ymax=82,
        tick label style={font=\scriptsize},
        ytick={20,30,...,80},
        ymajorgrids=true,
        grid style=dashed,
        bar width=2pt,
        legend style={at={(0.5,1.15)}, anchor=south, legend columns=5, font=\footnotesize},
        enlarge x limits=0.04,
        ylabel style={font=\small},
        title style={font=\small},
        nodes near coords,
        nodes near coords style={
            font=\fontsize{5}{6}\selectfont,
            rotate=90, 
            anchor=west, 
            /pgf/number format/fixed, 
            /pgf/number format/precision=1
        },
    ]

    \addplot[fill=colorBase, draw=none] coordinates {
        (afr-eng, 72.76) (amh-eng, 54.44) (arb-eng, 61.85) (arz-eng, 55.68) (fra-eng, 65.14) 
        (hau-eng, 52.74) (por-eng, 69.02) (som-eng, 49.08) (spa-eng, 56.69) (swh-eng, 60.61) 
        (wol-eng, 36.41) (yor-eng, 39.73) (zul-eng, 56.66) (lin-fra, 43.07) (wol-fra, 34.59)
    };

    \addplot[fill=colorFT, draw=none] coordinates {
        (afr-eng, 74.61) (amh-eng, 57.04) (arb-eng, 62.05) (arz-eng, 56.37) (fra-eng, 66.32) 
        (hau-eng, 54.78) (por-eng, 68.30) (som-eng, 49.57) (spa-eng, 56.83) (swh-eng, 63.41) 
        (wol-eng, 39.15) (yor-eng, 41.82) (zul-eng, 58.69) (lin-fra, 44.36) (wol-fra, 36.54)
    };

    \addplot[fill=color8, draw=none] coordinates {
        (afr-eng, 74.19) (amh-eng, 55.91) (arb-eng, 61.33) (arz-eng, 54.55) (fra-eng, 65.51) 
        (hau-eng, 54.15) (por-eng, 68.12) (som-eng, 49.37) (spa-eng, 56.50) (swh-eng, 62.36) 
        (wol-eng, 39.58) (yor-eng, 41.08) (zul-eng, 58.06) (lin-fra, 43.39) (wol-fra, 37.03)
    };

    \addplot[fill=color6, draw=none] coordinates {
        (afr-eng, 73.63) (amh-eng, 55.16) (arb-eng, 60.66) (arz-eng, 54.07) (fra-eng, 65.00) 
        (hau-eng, 53.66) (por-eng, 67.78) (som-eng, 48.73) (spa-eng, 56.35) (swh-eng, 61.96) 
        (wol-eng, 39.11) (yor-eng, 40.84) (zul-eng, 57.26) (lin-fra, 42.69) (wol-fra, 36.19)
    };

    \addplot[fill=color4, draw=none] coordinates {
        (afr-eng, 73.09) (amh-eng, 53.26) (arb-eng, 60.27) (arz-eng, 53.30) (fra-eng, 63.46) 
        (hau-eng, 52.86) (por-eng, 67.15) (som-eng, 47.44) (spa-eng, 55.75) (swh-eng, 61.11) 
        (wol-eng, 38.73) (yor-eng, 40.26) (zul-eng, 56.43) (lin-fra, 41.94) (wol-fra, 34.47)
    };

    \legend{Baseline, FT+KD, 8-dec, 6-dec, 4-dec}

    \end{axis}
    \end{tikzpicture}
    \caption{Translation performance (chrF++) from African languages to English/French.
    }
    \label{fig:afr-eng}
\end{figure*}

Furthermore, we thoroughly studied the effect of keeping all 12 encoder layers intact while iteratively removing different numbers of decoder layers. We experimented with three pruning configurations, removing 4, 6, or 8 decoder layers, resulting in models with 12 encoder layers and 8, 6, or 4 decoder layers, respectively. As illustrated in Figure \ref{fig:eng-afr} and Figure \ref{fig:afr-eng}, the effect of the number of decoder layers removed varies across language pairs, although removing up to 6 layers (50\%) yields similar or better performance compared to the NLLB-200 600M baseline, thanks to fine-tuning before and after pruning. Table \ref{tab:eval-all} elaborates further on the performance results in terms of both translation quality and inference speed.

\section{Conclusions and Future Work}

In this work, we presented AfriNLLB, lightweight models for African languages, that achieve over 20--50\% inference performance gains compared to their baseline NLLB-200 600M. We release models with various sizes to match different needs.

We have demonstrated that iterative layer pruning is an effective approach for model compression while retaining translation quality. The method relies on layer importance evaluation, followed by fine-tuning on a medium-sized dataset. This iterative layer pruning process reduces the model size and accelerates inference. We are open-sourcing AfriNLLB models and data. In addition, to ensure reproducibility, we are making all the processing and training code publicly available.

In future versions of AfriNLLB, we plan to add more languages. Research directions include investigating data augmentation approaches besides knowledge distillation, such as back-translation. Moreover, we plan to expand our approach to other architectures, such as autoregressive large language models and encoder-only models.

We hope that by releasing AfriNLLB models, training data, and code, we facilitate further research on African languages and support the African community worldwide.

\bibliography{paperafri,paperpile}

\appendix
\onecolumn

\newpage

\newpage
\begin{table*}[ht]
\centering

\begin{center}
    {\large \textbf{Performance Comparison: Layer Pruning Configurations}} \\
    \vspace{0.1cm}
    \small \textit{Translation quality (BLEU, chrF++, COMET) and efficiency (throughput, inference time) across baseline, fine-tuned, and pruned configurations with optional float16 (FP16) quantization}
\end{center}
\vspace{0.3cm}

\small
\setlength{\tabcolsep}{6pt}
\begin{tabular}{llcclccc|cc}
\toprule
\textbf{Lang} & \textbf{Model} & \textbf{Enc} & \textbf{Dec} & \textbf{Quant} & \textbf{BLEU ↑} & \textbf{chrF++ ↑} & \textbf{COMET ↑} & \textbf{Throughput ↑} & \textbf{Time ↓} \\
\midrule

\multirow{12}{*}{\textbf{xx-en}} 
& \multirow{2}{*}{NLLB} & \multirow{2}{*}{12} & \multirow{2}{*}{12} & -- & 33.81 & 56.22 & 71.11 & 1469.96 & 21.02 \\
& & & & FP16 & 33.80 & 56.22 & 71.13 & 2834.69 & 10.92 \\
\cline{2-10}\addlinespace[4pt]
& \multirow{2}{*}{NLLB + FT} & \multirow{2}{*}{12} & \multirow{2}{*}{12} & -- & 35.15 & 57.61 & 71.87 & 1530.94 & 20.39 \\
& & & & FP16 & 35.10 & 57.61 & 71.87 & 2808.90 & 11.15 \\
\cline{2-10}\addlinespace[4pt]
& \multirow{8}{*}{AfriNLLB} & \multirow{2}{*}{12} & \multirow{2}{*}{8} & -- & 34.01 & 56.98 & 71.20 & 1807.61 & 17.38 \\
& & & & FP16 & 34.05 & 56.99 & 71.19 & 3513.32 & 8.96 \\
\cline{3-10}\addlinespace[4pt]
& & \multirow{2}{*}{12} & \multirow{2}{*}{6} & -- & 33.35 & 56.48 & 70.79 & 2028.18 & 15.41 \\
& & & & FP16 & 33.32 & 56.45 & 70.79 & 4000.25 & 7.82 \\
\cline{3-10}\addlinespace[4pt]
& & \multirow{2}{*}{12} & \multirow{2}{*}{4} & -- & 32.03 & 55.62 & 69.71 & 2257.03 & 13.77 \\
& & & & FP16 & 32.01 & 55.60 & 69.71 & 4589.42 & 6.79 \\
\cline{3-10}\addlinespace[4pt]
& & \multirow{2}{*}{8} & \multirow{2}{*}{8} & -- & 30.89 & 54.32 & 68.08 & 1852.13 & 17.05 \\
& & & & FP16 & 30.86 & 54.30 & 68.08 & 3550.50 & 8.91 \\

\midrule

\multirow{12}{*}{\textbf{en-xx}} 
& \multirow{2}{*}{NLLB} & \multirow{2}{*}{12} & \multirow{2}{*}{12} & -- & 22.70 & 47.89 & 69.36 & 1530.10 & 28.09 \\
& & & & FP16 & 22.68 & 47.88 & 69.38 & 2898.38 & 15.33 \\
\cline{2-10}\addlinespace[4pt]
& \multirow{2}{*}{NLLB + FT} & \multirow{2}{*}{12} & \multirow{2}{*}{12} & -- & 24.28 & 49.97 & 70.91 & 1610.23 & 26.98 \\
& & & & FP16 & 24.14 & 49.84 & 70.90 & 2811.34 & 18.82 \\
\cline{2-10}\addlinespace[4pt]
& \multirow{8}{*}{AfriNLLB} & \multirow{2}{*}{12} & \multirow{2}{*}{8} & -- & 24.17 & 50.05 & 70.37 & 1946.61 & 22.51 \\
& & & & FP16 & 24.15 & 50.06 & 70.41 & 3732.72 & 11.98 \\
\cline{3-10}\addlinespace[4pt]
& & \multirow{2}{*}{12} & \multirow{2}{*}{6} & -- & 23.48 & 49.34 & 68.98 & 2265.87 & 18.50 \\
& & & & FP16 & 23.49 & 49.35 & 69.00 & 4428.68 & 9.65 \\
\cline{3-10}\addlinespace[4pt]
& & \multirow{2}{*}{12} & \multirow{2}{*}{4} & -- & 21.77 & 47.80 & 65.68 & 2489.35 & 17.31 \\
& & & & FP16 & 21.77 & 47.81 & 65.68 & 4954.62 & 9.09 \\
\cline{3-10}\addlinespace[4pt]
& & \multirow{2}{*}{8} & \multirow{2}{*}{8} & -- & 23.59 & 49.64 & 69.90 & 2015.53 & 21.34 \\
& & & & FP16 & 23.58 & 49.63 & 69.88 & 3851.13 & 11.34 \\

\midrule \midrule

\multirow{12}{*}{\textbf{xx-fr}} 
& \multirow{2}{*}{NLLB} & \multirow{2}{*}{12} & \multirow{2}{*}{12} & -- & 16.41 & 38.83 & 17.34 & 1475.48 & 26.46 \\
& & & & FP16 & 16.33 & 38.83 & 17.23 & 2850.66 & 13.71 \\
\cline{2-10}\addlinespace[4pt]
& \multirow{2}{*}{NLLB + FT} & \multirow{2}{*}{12} & \multirow{2}{*}{12} & -- & 17.91 & 40.45 & 18.47 & 1524.32 & 26.12 \\
& & & & FP16 & 17.83 & 40.42 & 18.37 & 2749.45 & 14.68 \\
\cline{2-10}\addlinespace[4pt]
& \multirow{8}{*}{AfriNLLB} & \multirow{2}{*}{12} & \multirow{2}{*}{8} & -- & 17.43 & 40.21 & 14.52 & 1845.09 & 21.61 \\
& & & & FP16 & 17.38 & 40.18 & 14.53 & 3569.23 & 11.17 \\
\cline{3-10}\addlinespace[4pt]
& & \multirow{2}{*}{12} & \multirow{2}{*}{6} & -- & 16.52 & 39.44 & 11.78 & 2044.27 & 19.21 \\
& & & & FP16 & 16.54 & 39.42 & 11.68 & 3953.51 & 9.92 \\
\cline{3-10}\addlinespace[4pt]
& & \multirow{2}{*}{12} & \multirow{2}{*}{4} & -- & 14.96 & 38.21 & 5.67 & 2340.99 & 16.77 \\
& & & & FP16 & 14.90 & 38.17 & 5.71 & 4766.12 & 8.24 \\
\cline{3-10}\addlinespace[4pt]
& & \multirow{2}{*}{8} & \multirow{2}{*}{8} & -- & 14.42 & 37.05 & 3.14 & 1866.26 & 21.84 \\
& & & & FP16 & 14.34 & 36.97 & 3.14 & 3448.51 & 11.83 \\

\midrule

\multirow{12}{*}{\textbf{fr-xx}} 
& \multirow{2}{*}{NLLB} & \multirow{2}{*}{12} & \multirow{2}{*}{12} & -- & 9.44 & 33.42 & 19.25 & 1047.18 & 49.92 \\
& & & & FP16 & 9.52 & 33.40 & 19.38 & 1920.41 & 29.05 \\
\cline{2-10}\addlinespace[4pt]
& \multirow{2}{*}{NLLB + FT} & \multirow{2}{*}{12} & \multirow{2}{*}{12} & -- & 10.98 & 35.68 & 21.33 & 1081.84 & 51.56 \\
& & & & FP16 & 10.48 & 35.05 & 21.49 & 1700.25 & 51.31 \\
\cline{2-10}\addlinespace[4pt]
& \multirow{8}{*}{AfriNLLB} & \multirow{2}{*}{12} & \multirow{2}{*}{8} & -- & 10.20 & 35.21 & 20.04 & 1261.66 & 49.91 \\
& & & & FP16 & 10.11 & 35.13 & 20.03 & 2313.85 & 31.15 \\
\cline{3-10}\addlinespace[4pt]
& & \multirow{2}{*}{12} & \multirow{2}{*}{6} & -- & 10.07 & 35.14 & 19.83 & 1416.33 & 30.89 \\
& & & & FP16 & 9.99 & 35.08 & 19.78 & 2465.60 & 18.68 \\
\cline{3-10}\addlinespace[4pt]
& & \multirow{2}{*}{12} & \multirow{2}{*}{4} & -- & 7.57 & 32.42 & 14.16 & 1207.06 & 38.75 \\
& & & & FP16 & 7.57 & 32.38 & 14.29 & 2069.52 & 23.25 \\
\cline{3-10}\addlinespace[4pt]
& & \multirow{2}{*}{8} & \multirow{2}{*}{8} & -- & 9.75 & 35.23 & 20.05 & 1222.83 & 45.33 \\
& & & & FP16 & 9.84 & 35.31 & 20.11 & 2186.73 & 25.97 \\

\bottomrule
\end{tabular}
\caption{Comprehensive performance evaluation across translation directions. AfriNLLB models use various encoder-decoder layer configurations (12-8, 12-6, 12-4, 8-8) with and without float16 quantization.}
\label{tab:eval-all}
\end{table*}

\clearpage
\newpage
\begin{center}
    {\large \textbf{Datasets Sources and Sizes}} \\
    \vspace{0.1cm}
    \small \textit{Names, sources, and sizes of our training datasets before and after filtering for each language pair}
\end{center}

\begin{table*}[htbp]
\centering

\begin{adjustbox}{max width=\textwidth,center}
\scriptsize
\begin{tabular}{l|c|c|c|c|c|c|c|c|c|c|c|c|c|c|c}
\toprule
\textbf{Dataset} & \textbf{fra-eng} & \textbf{spa-eng} & \textbf{por-eng} & \textbf{arb-eng} & \textbf{swh-eng} & \textbf{amh-eng} & \textbf{som-eng} & \textbf{hau-eng} & \textbf{yor-eng} & \textbf{zul-eng} & \textbf{afr-eng} & \textbf{arz-eng} & \textbf{wol-fra} & \textbf{wol-eng} & \textbf{lin-fra} \\
\midrule
\multicolumn{16}{c}{\textit{OPUS Datasets}} \\
\midrule
Tatoeba \cite{Tiedemann2020-Tatoeba} & -- & -- & -- & -- & -- & 213/188 & 9/5 & 259/183 & 423/421 & 72/170 & 2.4K/2.1K & 6.5K/1.3K & -- & -- & 555/120 \\
translatewiki & -- & -- & -- & -- & -- & -- & -- & -- & -- & 6.2K/244 & -- & 111K/23K & 1.7K/243 & -- & -- \\
wikimedia & 1.4M/1.1M & -- & 822K/610K & 621K/374K & 16.3K/11.3K & 942/425 & 1.1K/718 & 190K/121K & 12.5K/4.8K & 9.3K/5.5K & 78.5K/66.5K & -- & 690/169 & 21/5 & -- \\
GNOME & -- & -- & 21.2K/15.3K & 150/41 & 40/43 & 57.1K/26.9K & 753/1.1K & 5.5K/110 & 1K/590 & 4.5K/7.7K & 12.7K/27.8K & -- & -- & -- & -- \\
Ubuntu & -- & -- & -- & 6K/2.5K & -- & -- & -- & -- & 141/0 & -- & -- & -- & 220/38 & 222/26 & -- \\
GlobalVoices & -- & -- & -- & -- & 32.3K/26.9K & 1.8K/1.2K & -- & -- & 136/61 & -- & -- & -- & -- & -- & -- \\
bible-uedin \cite{christodouloupoulos2015uedin} & -- & -- & -- & 62.2K/16.3K & -- & 6.1K/46.6K & 6.2K/49.5K & -- & -- & -- & 62.1K/50.6K & -- & 7.9K/648 & 15.8K/2.6K & -- \\
NeuLab-TedTalks & 212K/185K & 215K/190K & 81.2K/55K & -- & -- & -- & -- & -- & -- & -- & -- & -- & -- & -- & -- \\
EMEA & -- & 1.1M/235K & 1.1M/223K & -- & -- & -- & -- & -- & -- & -- & -- & -- & -- & -- & -- \\
EUbookshop & -- & -- & 4.2M/610K & -- & 1.7K/110 & -- & -- & -- & -- & -- & -- & -- & -- & -- & -- \\
ELRC-wiki\_health & 4.4K/3.7K & -- & -- & 15.1K/14.4K & 608/501 & -- & -- & -- & -- & -- & 404/312 & -- & -- & -- & -- \\
News-Commentary & 156K/125K & -- & -- & -- & -- & -- & -- & -- & -- & -- & -- & -- & -- & -- & -- \\
JRC-Acquis \cite{steinberger2006jrc} & 814K/65.3K & 806K/398K & -- & -- & -- & -- & -- & -- & -- & -- & -- & -- & -- & -- & -- \\
TED2020 & -- & -- & -- & 408K/341K & 9.7K/80.8K & 1K/1.7K & 2K/1.3K & 27/21 & -- & -- & 2.3K/1.8K & -- & -- & -- & -- \\
KDE4 & -- & -- & -- & 116K/25.6K & -- & -- & -- & 149/66 & -- & -- & 64.3K/29.8K & -- & -- & -- & -- \\
ELRC-EMEA & -- & 777K/614K & -- & -- & -- & -- & -- & -- & -- & -- & -- & -- & -- & -- & -- \\
Books & -- & 93.5K/63.4K & -- & -- & -- & -- & -- & -- & -- & -- & -- & -- & -- & -- & -- \\
Tanzil & -- & -- & -- & -- & 138K/96.7K & 93.5K/50.5K & 93.8K/10.5K & 128K/63.4K & -- & -- & -- & -- & -- & -- & -- \\
OpenSubtitles & -- & -- & -- & -- & 94.6K/95.8K & 3K/1.6K & 531/446 & -- & -- & -- & 969K/11.8K & -- & -- & -- & -- \\
TICO-19 \cite{Anastasopoulos2020-TICO-19} & -- & -- & -- & -- & 3.1K/2.8K & 3.1K/3.1K & 3.1K/1.2K & 3.1K/2.1K & -- & 3.1K/2.3K & -- & -- & -- & -- & 2.9K/544 \\
ELRC\_2922 & -- & -- & -- & -- & 607/498 & -- & -- & -- & -- & -- & 403/310 & -- & -- & -- & -- \\
ELRC-3073-wiki\_health & -- & -- & -- & -- & 608/501 & -- & -- & -- & -- & -- & -- & -- & -- & -- & -- \\
infopankki & -- & -- & -- & -- & -- & -- & 47.2K/89.8K & -- & -- & -- & -- & -- & -- & -- & -- \\
QED & -- & -- & -- & -- & -- & -- & -- & -- & -- & -- & 28.8K/17.5K & -- & -- & -- & -- \\
SPC & -- & -- & -- & -- & -- & -- & -- & -- & -- & -- & 57.4K/47.3K & -- & -- & -- & -- \\
ELRC-monumentos & -- & -- & -- & -- & -- & -- & -- & -- & -- & -- & 54/41 & -- & -- & -- & -- \\
ELRC-Museus & -- & -- & -- & -- & -- & -- & -- & -- & -- & -- & 32/0 & -- & -- & -- & -- \\
\midrule
\multicolumn{16}{c}{\textit{HuggingFace Datasets}} \\
\midrule
smol \cite{caswell2025smol} & -- & -- & -- & -- & 863/719 & 863/712 & 862/551 & 863/548 & 863/153 & 863/552 & 863/610 & -- & -- & 7.4K/570 & -- \\
mafand \cite{adelani2022mafand} & -- & -- & -- & -- & 34.4K/29.9K & 1.9K/1.4K & -- & 5.9K/4.4K & 6.6K/4K & 3.5K/2K & -- & -- & -- & -- & -- \\
mafand-dev & -- & -- & -- & -- & -- & -- & -- & 1.3K/971 & 6.6K/4K & 1.2K/636 & -- & -- & -- & -- & -- \\
mafand-test & -- & -- & -- & -- & -- & -- & -- & 1.5K/1.2K & 6.6K/4K & 998/596 & -- & -- & -- & -- & -- \\
Pontoon-Translations & -- & -- & -- & -- & 6.1K/2.8K & -- & -- & 3.2K/1.2K & 4.4K/553 & 3.3K/735 & 13.1K/2.7K & -- & -- & 6.8K/802 & -- \\
Weblate-Translations & -- & -- & -- & -- & 17.2K/7.2K & 8K/2.1K & 1.6K/310 & 143/54 & 164/533 & 66/53 & 23.2K/1.8K & -- & -- & -- & -- \\
ntrex \cite{federmann2022ntrex} & -- & -- & -- & -- & 2K/1.7K & 2K/1.5K & 2K/1.9K & 2K/1.2K & 2K/602 & 2K/1K & 2K/1.8K & -- & -- & -- & -- \\
AfriDocMT-health \cite{adelani2025afridocmt} & -- & -- & -- & -- & 7K/6.6K & 7K/6.6K & -- & 7K/5.9K & 7K/3.7K & 7K/5.4K & -- & -- & -- & -- & -- \\
AfriDocMT-doc\_health & -- & -- & -- & -- & 240/4 & 240/6 & -- & -- & -- & -- & -- & -- & -- & -- & -- \\
AfriDocMT-doc\_health\_2 & -- & -- & -- & -- & 540/96 & 540/104 & -- & -- & -- & -- & -- & -- & -- & -- & -- \\
AfriDocMT-doc\_health\_5 & -- & -- & -- & -- & 1.5K/1.5K & 1.5K/1.5K & -- & -- & -- & -- & -- & -- & -- & -- & -- \\
AfriDocMT-doc\_health\_10 & -- & -- & -- & -- & 812/366 & 812/440 & -- & -- & -- & -- & -- & -- & -- & -- & -- \\
quran\_multilingual & -- & -- & -- & -- & -- & 6.2K/1K & 6.2K/740 & 6.2K/3.8K & 6.2K/1K & -- & -- & -- & -- & -- & -- \\
Nazimali-Quran & -- & -- & -- & -- & 6.2K/5K & 6.2K/3.7K & 6.2K/5K & -- & -- & -- & -- & -- & -- & -- & -- \\
OPUS-100 \cite{zhang2020opus100} & -- & -- & -- & -- & -- & -- & -- & -- & 10.4K/2.3K & -- & -- & -- & -- & -- & -- \\
OPUS-100-dev & -- & -- & -- & -- & -- & -- & -- & -- & 10.4K/2.3K & -- & -- & -- & -- & -- & -- \\
OPUS-100-test & -- & -- & -- & -- & -- & -- & -- & -- & 10.4K/2.3K & -- & -- & -- & -- & -- & -- \\
menyo20k\_mt-train \cite{adelani2021menyo} & -- & -- & -- & -- & -- & -- & -- & -- & 10.1K/4.6K & -- & -- & -- & -- & -- & -- \\
menyo20k\_mt-dev & -- & -- & -- & -- & -- & -- & -- & -- & 3.4K/1.4K & -- & -- & -- & -- & -- & -- \\
menyo20k\_mt-test & -- & -- & -- & -- & -- & -- & -- & -- & 6.6K/3.7K & -- & -- & -- & -- & -- & -- \\
yoruba\_audio\_trans & -- & -- & -- & -- & -- & -- & -- & -- & 9.2K/1.9K & -- & -- & -- & -- & -- & -- \\
arz-en-parallel & -- & -- & -- & -- & -- & -- & -- & -- & -- & -- & -- & 25K/22.6K & -- & -- & -- \\
news-comm-eng-arz \cite{Moslem2025-IterativePruning} & -- & -- & -- & -- & -- & -- & -- & -- & -- & -- & -- & 832K/83.3K & -- & -- & -- \\
mteb/tatoeba-bitext \cite{enevoldsen2025mmteb} & -- & -- & -- & -- & -- & -- & -- & -- & -- & -- & -- & 8.9K/2.9K & -- & -- & -- \\
fr-wolof-trans-gs & -- & -- & -- & -- & -- & -- & -- & -- & -- & -- & -- & -- & 10.4K/1.6K & -- & -- \\
wolof\_en\_fr & -- & -- & -- & -- & -- & -- & -- & -- & -- & -- & -- & -- & 26.6K/6.5K & 26.6K/7.6K & -- \\
english\_wolof\_trans & -- & -- & -- & -- & -- & -- & -- & -- & -- & -- & -- & -- & -- & 84.7K/17.2K & -- \\
comet\_score\_en\_wo & -- & -- & -- & -- & -- & -- & -- & -- & -- & -- & -- & -- & -- & 7.5K/4K & -- \\
wolof\_en\_bible & -- & -- & -- & -- & -- & -- & -- & -- & -- & -- & -- & -- & -- & 13.4K/2.2K & -- \\
MultiUN \cite{eisele2010multiun} & -- & -- & -- & 9.8M/9.8M & -- & -- & -- & -- & -- & -- & -- & -- & -- & -- & -- \\
ted\_talks\_iwslt-14 \cite{cettolo2012TED} & -- & -- & -- & -- & 52/42 & -- & -- & -- & -- & -- & -- & -- & -- & -- & -- \\
ted\_talks\_iwslt-15 & -- & -- & -- & -- & 68/53 & -- & -- & -- & -- & -- & -- & -- & -- & -- & -- \\
ted\_talks\_iwslt & -- & -- & -- & -- & -- & -- & 188/730 & -- & -- & -- & -- & -- & -- & -- & -- \\
WMT24pp \cite{deutsch2025wmt24} & -- & -- & -- & -- & 998/691 & -- & -- & -- & -- & -- & -- & -- & -- & -- & -- \\
sunbird-salt \cite{kumbuga2024salt} & -- & -- & -- & -- & 24.9K/23.1K & -- & -- & -- & -- & -- & -- & -- & -- & -- & -- \\
HausaVG \cite{abdulmumin2022HausaVG} & -- & -- & -- & -- & -- & -- & -- & 28.9K/7.9K & -- & -- & -- & -- & -- & -- & -- \\
polynews-parallel \cite{iana2023polynews} & -- & -- & -- & -- & -- & -- & -- & 5.7K/4.4K & -- & 3.4K/2K & -- & -- & -- & -- & -- \\
Quran & -- & -- & -- & -- & -- & -- & -- & 6.2K/3.7K & -- & -- & -- & -- & -- & -- & -- \\
hf-spc & -- & -- & -- & -- & -- & -- & -- & -- & -- & -- & 57.4K/47.4K & -- & -- & -- & -- \\
lingvanex\_test & -- & -- & -- & -- & -- & -- & -- & -- & -- & -- & 1.1K/649 & -- & -- & -- & -- \\
subscene & -- & -- & -- & -- & -- & -- & 900/0 & -- & -- & -- & -- & -- & -- & -- & -- \\
opus\_infopankki & -- & -- & -- & -- & -- & -- & 47.2K/89.8K & -- & -- & -- & -- & -- & -- & -- & -- \\
\midrule
\multicolumn{16}{c}{\textit{other sources}} \\
\midrule
ArzEn-MultiGenre \cite{alsabbagh2024arz} & -- & -- & -- & -- & -- & -- & -- & -- & -- & -- & -- & 25K/6.6K & -- & -- & -- \\
ethiopian-legal & -- & -- & -- & -- & -- & 5.4K/3.7K & -- & -- & -- & -- & -- & -- & -- & -- & -- \\
ethiopian-history & -- & -- & -- & -- & -- & 1.3K/737 & -- & -- & -- & -- & -- & -- & -- & -- & -- \\
ethiopian-news & -- & -- & -- & -- & -- & 5.4K/1.1K & -- & -- & -- & -- & -- & -- & -- & -- & -- \\
ethiopian-ebible & -- & -- & -- & -- & -- & 6.5K/3.8K & -- & -- & -- & -- & -- & -- & -- & -- & -- \\
ethiopian-ethio\_bible & -- & -- & -- & -- & -- & 11.7K/5.7K & -- & -- & -- & -- & -- & -- & -- & -- & -- \\
ethiopian-jw\_bible & -- & -- & -- & -- & -- & 31.1K/25.2K & -- & -- & -- & -- & -- & -- & -- & -- & -- \\
ethiopian-jw\_daily & -- & -- & -- & -- & -- & 4.7K/4.3K & -- & -- & -- & -- & -- & -- & -- & -- & -- \\
horn-mt & -- & -- & -- & -- & -- & 2K/2K & -- & -- & -- & -- & -- & -- & -- & -- & -- \\
mt-eval-am-amen & -- & -- & -- & -- & -- & 997/712 & -- & -- & -- & -- & -- & -- & -- & -- & -- \\
mt-eval-am-enam & -- & -- & -- & -- & -- & 1.9K/1.4K & -- & -- & -- & -- & -- & -- & -- & -- & -- \\
ukuxhumana & -- & -- & -- & -- & -- & -- & -- & -- & -- & 26.7K/13.8K & -- & -- & -- & -- & -- \\
zenodo-training & -- & -- & -- & -- & -- & -- & -- & -- & -- & 4.7K/2.6K & -- & -- & -- & -- & -- \\
zenodo-eval & -- & -- & -- & -- & -- & -- & -- & -- & -- & 998/596 & -- & -- & -- & -- & -- \\
Gamayun-fr-ln & -- & -- & -- & -- & -- & -- & -- & -- & -- & -- & -- & -- & -- & -- & 5K/1.4K \\
\midrule
\midrule
\textbf{Total (Origin)} & 2.6M & 3M & 6.2M & 11M & 399K & 319K & 275K & 398K & 124K & 113K & 619K & 234K & 47.5K & 162K & 8.1K \\
\textbf{After Filter} & 1.5M & 1.5M & 1.5M & 10.5M & 287K & 157K & 87.5K & 222K & 34.7K & 38.5K & 174K & 85.9K & 9.2K & 35K & 2K \\
\textbf{After Dedup} & 1.48M & 1.32M & 1.4M & 1.42M & 181K & 85K & 87.5K & 156K & 22.6K & 33.2K & 174K & 84.2K & 9.1K & 31.2K & 1.9K \\
\bottomrule
\end{tabular}
\end{adjustbox}

\vspace{0.3cm}

\caption{Dataset statistics for all language pairs. Values shown as Original/Final (K=thousand, M=million), and ``--'' indicates dataset not used.}
\label{tab:data-details}

\end{table*}

\newpage
\setlength{\LTleft}{0pt}
\setlength{\LTright}{0pt}

\begin{center}
    {\large \textbf{Performance Comparison: NLLB-200 600M vs. AfriNLLB 548M Models}} \\
    \vspace{0.1cm}
    \small \textit{Comparison of NLLB-200 600M baseline, Pruned (Iterative) + Fine-tuned, and Pruned (Iterative) + Fine-tuned (float16 quantization) across BLEU, chrF++, COMET, and Output Throughput (output tokens/second)}
\end{center}

\tiny
\setlength{\tabcolsep}{1.5pt}
\begin{longtable}{l|cccc|cccr|cccr}
\toprule
\textbf{Lang} & \multicolumn{4}{c|}{\textbf{NLLB 600M (baseline)}} & \multicolumn{4}{c|}{\textbf{AfriNLLB 548M (Iterative FT)}} & \multicolumn{4}{c}{\textbf{AfriNLLB 548M (Iterative FT FP16)}} \\
\textbf{Pair} & \textbf{BLEU} & \textbf{chrF++} & \textbf{COMET} & \textbf{Throughput} & \textbf{BLEU} & \textbf{chrF++} & \textbf{COMET} & \textbf{Throughput} & \textbf{BLEU} & \textbf{chrF++} & \textbf{COMET} & \textbf{Throughput} \\
\midrule
\endfirsthead

en-af & 35.82 & 61.76 & 75.10 & 1,672 & 41.17 \up{14.9\%} & 66.23 \up{7.2\%} & 75.86 \up{1.0\%} & 2,147 \up{28.4\%} & 41.08 \up{14.7\%} & 66.20 \up{7.2\%} & 75.85 \up{1.0\%} & 4,222 \up{152.5\%} \\
af-en & 54.18 & 72.76 & 74.02 & 1,484 & 56.09 \up{3.5\%} & 74.19 \up{2.0\%} & 74.28 \up{0.4\%} & 1,912 \up{28.8\%} & 56.08 \up{3.5\%} & 74.17 \up{1.9\%} & 74.30 \up{0.4\%} & 3,758 \up{153.3\%} \\
\midrule
en-am & 12.12 & 36.96 & 69.24 & 1,407 & 12.82 \up{5.8\%} & 38.37 \up{3.8\%} & 70.72 \up{2.1\%} & 1,816 \up{29.1\%} & 12.86 \up{6.1\%} & 38.42 \up{4.0\%} & 70.85 \up{2.3\%} & 3,137 \up{123.0\%} \\
am-en & 30.02 & 54.44 & 67.88 & 1,542 & 31.56 \up{5.1\%} & 55.91 \up{2.7\%} & 67.22 \down{1.0\%} & 1,797 \up{16.5\%} & 31.57 \up{5.2\%} & 55.89 \up{2.7\%} & 67.20 \down{1.0\%} & 3,454 \up{124.0\%} \\
\midrule
en-ar & 22.86 & 51.30 & 84.78 & 1,648 & 24.16 \up{5.7\%} & 52.51 \up{2.4\%} & 84.94 \up{0.2\%} & 2,104 \up{27.7\%} & 24.18 \up{5.8\%} & 52.50 \up{2.3\%} & 84.92 \up{0.2\%} & 4,111 \up{149.5\%} \\
ar-en & 39.00 & 61.85 & 85.99 & 1,415 & 37.00 \down{5.1\%} & 61.33 \down{0.8\%} & 85.74 \down{0.3\%} & 1,732 \up{22.4\%} & 37.11 \down{4.8\%} & 61.33 \down{0.8\%} & 85.73 \down{0.3\%} & 3,345 \up{136.4\%} \\
\midrule
en-arz & 11.87 & 40.81 & 80.16 & 1,525 & 14.94 \up{25.9\%} & 44.43 \up{8.9\%} & 81.98 \up{2.3\%} & 2,063 \up{35.3\%} & 14.95 \up{25.9\%} & 44.46 \up{8.9\%} & 81.99 \up{2.3\%} & 4,042 \up{165.0\%} \\
arz-en & 30.64 & 55.68 & 82.65 & 1,382 & 28.69 \down{6.4\%} & 54.55 \down{2.0\%} & 82.11 \down{0.7\%} & 1,753 \up{26.8\%} & 28.77 \down{6.1\%} & 54.58 \down{2.0\%} & 82.11 \down{0.7\%} & 3,422 \up{147.6\%} \\
\midrule
en-es & 26.71 & 52.59 & 85.30 & 1,696 & 24.78 \down{7.2\%} & 51.40 \down{2.3\%} & 84.38 \down{1.1\%} & 2,152 \up{26.9\%} & 24.71 \down{7.5\%} & 51.37 \down{2.3\%} & 84.40 \down{1.1\%} & 4,210 \up{148.2\%} \\
es-en & 29.91 & 56.69 & 86.11 & 1,571 & 27.99 \down{6.4\%} & 56.50 \down{0.3\%} & 86.05 \down{0.1\%} & 1,903 \up{21.1\%} & 28.00 \down{6.4\%} & 56.46 \down{0.4\%} & 86.03 \down{0.1\%} & 3,738 \up{138.0\%} \\
\midrule
en-fr & 46.70 & 66.61 & 86.78 & 1,700 & 46.16 \down{1.2\%} & 66.99 \up{0.6\%} & 86.25 \down{0.6\%} & 2,118 \up{24.6\%} & 46.25 \down{1.0\%} & 67.05 \up{0.7\%} & 86.26 \down{0.6\%} & 4,161 \up{144.8\%} \\
fr-en & 43.15 & 65.14 & 88.19 & 1,454 & 41.73 \down{3.3\%} & 65.51 \up{0.6\%} & 88.17 \down{0.0\%} & 1,819 \up{25.1\%} & 41.88 \down{2.9\%} & 65.56 \up{0.6\%} & 88.18 \down{0.0\%} & 3,537 \up{143.3\%} \\
\midrule
en-ha & 23.69 & 48.99 & 63.93 & 1,596 & 27.64 \up{16.7\%} & 53.21 \up{8.6\%} & 65.01 \up{1.7\%} & 1,894 \up{18.7\%} & 27.61 \up{16.5\%} & 53.22 \up{8.6\%} & 65.12 \up{1.9\%} & 3,583 \up{124.6\%} \\
ha-en & 31.06 & 52.74 & 65.97 & 1,514 & 32.36 \up{4.2\%} & 54.15 \up{2.7\%} & 66.37 \up{0.6\%} & 1,907 \up{26.0\%} & 32.48 \up{4.6\%} & 54.22 \up{2.8\%} & 66.31 \up{0.5\%} & 3,754 \up{147.9\%} \\
\midrule
fr-ln & 15.15 & 45.00 & 35.84 & 1,419 & 15.88 \up{4.8\%} & 44.97 \down{0.1\%} & 34.07 \down{4.9\%} & 1,789 \up{26.1\%} & 15.72 \up{3.8\%} & 44.87 \down{0.3\%} & 33.85 \down{5.6\%} & 3,498 \up{146.5\%} \\
ln-fr & 19.85 & 43.07 & 36.61 & 1,559 & 20.06 \up{1.1\%} & 43.39 \up{0.7\%} & 30.12 \down{17.7\%} & 1,906 \up{22.3\%} & 20.00 \up{0.8\%} & 43.35 \up{0.7\%} & 30.09 \down{17.8\%} & 3,663 \up{135.0\%} \\
\midrule
en-pt & 46.45 & 67.17 & 88.56 & 1,726 & 42.72 \down{8.0\%} & 65.33 \down{2.7\%} & 87.74 \down{0.9\%} & 2,150 \up{24.6\%} & 42.56 \down{8.4\%} & 65.23 \down{2.9\%} & 87.72 \down{0.9\%} & 4,224 \up{144.7\%} \\
pt-en & 48.08 & 69.02 & 88.95 & 1,580 & 46.48 \down{3.3\%} & 68.12 \down{1.3\%} & 88.57 \down{0.4\%} & 1,949 \up{23.4\%} & 46.62 \down{3.0\%} & 68.15 \down{1.3\%} & 88.59 \down{0.4\%} & 3,832 \up{142.5\%} \\
\midrule
en-so & 11.38 & 41.45 & 61.63 & 1,600 & 11.05 \down{2.9\%} & 40.98 \down{1.1\%} & 57.92 \down{6.0\%} & 1,799 \up{12.4\%} & 11.03 \down{3.1\%} & 40.98 \down{1.1\%} & 58.00 \down{5.9\%} & 3,400 \up{112.5\%} \\
so-en & 26.20 & 49.08 & 61.31 & 1,371 & 26.36 \up{0.6\%} & 49.37 \up{0.6\%} & 60.14 \down{1.9\%} & 1,718 \up{25.3\%} & 26.42 \up{0.8\%} & 49.41 \up{0.7\%} & 60.10 \down{2.0\%} & 3,328 \up{142.7\%} \\
\midrule
en-sw & 31.50 & 57.68 & 70.14 & 1,780 & 36.77 \up{16.7\%} & 62.23 \up{7.9\%} & 71.72 \up{2.3\%} & 2,138 \up{20.1\%} & 36.78 \up{16.8\%} & 62.25 \up{7.9\%} & 71.73 \up{2.3\%} & 4,168 \up{134.2\%} \\
sw-en & 39.47 & 60.61 & 70.12 & 1,672 & 41.40 \up{4.9\%} & 62.36 \up{2.9\%} & 70.38 \up{0.4\%} & 1,974 \up{18.1\%} & 41.41 \up{4.9\%} & 62.40 \up{3.0\%} & 70.42 \up{0.4\%} & 3,875 \up{131.8\%} \\
\midrule
en-wo & 5.06 & 23.56 & 17.27 & 923 & 6.97 \up{37.7\%} & 28.63 \up{21.5\%} & 22.65 \up{31.2\%} & 992 \up{7.5\%} & 6.99 \up{38.1\%} & 28.75 \up{22.0\%} & 22.75 \up{31.7\%} & 1,678 \up{81.8\%} \\
wo-en & 14.90 & 36.41 & 36.16 & 1,403 & 17.11 \up{14.8\%} & 39.58 \up{8.7\%} & 38.37 \up{6.1\%} & 1,596 \up{13.8\%} & 16.98 \up{14.0\%} & 39.53 \up{8.6\%} & 38.38 \up{6.1\%} & 3,011 \up{114.6\%} \\
\midrule
wo-fr & 12.96 & 34.59 & -1.93 & 1,392 & 14.79 \up{14.1\%} & 37.03 \up{7.1\%} & -1.09 \up{43.5\%} & 1,784 \up{28.2\%} & 14.76 \up{13.9\%} & 37.00 \up{7.0\%} & -1.03 \up{46.6\%} & 3,475 \up{149.6\%} \\
fr-wo & 3.73 & 21.84 & 2.66 & 676 & 4.52 \up{21.2\%} & 25.45 \up{16.5\%} & 6.01 \up{125.9\%} & 735 \up{8.7\%} & 4.49 \up{20.4\%} & 25.38 \up{16.2\%} & 6.21 \up{133.5\%} & 1,130 \up{67.2\%} \\
\midrule
en-yo & 4.32 & 22.87 & 51.89 & 1,002 & 8.03 \up{85.9\%} & 29.20 \up{27.7\%} & 59.51 \up{14.7\%} & 1,820 \up{81.6\%} & 8.05 \up{86.3\%} & 29.19 \up{27.6\%} & 59.63 \up{14.9\%} & 3,437 \up{243.1\%} \\
yo-en & 17.61 & 39.73 & 49.68 & 1,295 & 18.62 \up{5.7\%} & 41.08 \up{3.4\%} & 50.45 \up{1.5\%} & 1,590 \up{22.8\%} & 18.72 \up{6.3\%} & 41.18 \up{3.6\%} & 50.35 \up{1.3\%} & 3,014 \up{132.7\%} \\
\midrule
en-zu & 16.68 & 50.78 & 66.95 & 1,616 & 16.98 \up{1.8\%} & 51.19 \up{0.8\%} & 66.12 \down{1.2\%} & 2,116 \up{30.9\%} & 16.92 \up{1.4\%} & 51.14 \up{0.7\%} & 66.07 \down{1.3\%} & 4,152 \up{157.0\%} \\
zu-en & 35.32 & 56.66 & 67.45 & 1,427 & 36.77 \up{4.1\%} & 58.06 \up{2.5\%} & 67.80 \up{0.5\%} & 1,848 \up{29.5\%} & 36.56 \up{3.5\%} & 57.96 \up{2.3\%} & 67.76 \up{0.5\%} & 3,606 \up{152.7\%} \\
\midrule
\midrule
\textbf{Average} & \textbf{26.21} & \textbf{49.93} & \textbf{63.31} & \textbf{1468.2} & \textbf{27.05} \up{3.2\%} & \textbf{51.41} \up{3.0\%} & \textbf{63.65} \up{0.54\%} & \textbf{1833.95} \up{24.9\%} & \textbf{27.05} \up{3.2\%} & \textbf{51.41} \up{3.0\%} & \textbf{63.66} \up{0.55\%} & \textbf{3532.15} \up{140.57\%} \\
\bottomrule
\caption{Detailed evaluation of AfriNLLB models for each language direction. Overall, the compressed models achieve comparable or improved translation quality while yielding significant inference throughput gains over the baseline NLLB-200 600M.}
\label{tab:eval-details}
\end{longtable}

\end{document}